\documentclass{article}
\usepackage{iclr2022_workshop,times}
\usepackage{graphicx}

\usepackage{amsmath,amsfonts,bm}

\def\eqref#1{equation~\ref{#1}}

\def\1{\bm{1}}

\DeclareMathAlphabet{\mathsfit}{\encodingdefault}{\sfdefault}{m}{sl}
\SetMathAlphabet{\mathsfit}{bold}{\encodingdefault}{\sfdefault}{bx}{n}

\usepackage{hyperref}
\usepackage{xurl}

\title{Manifold-aligned Neighbor Embedding}

\author{Mohammad Tariqul Islam, Jason W. Fleischer\thanks{Corresponding author.} \\
Department of Electrical and Computer Engineering\\
Princeton University\\
Princeton, NJ 08544 \\
\texttt{\{mtislam,jasonf\}@princeton.edu}
}

\iclrfinalcopy 
\begin{document}

\maketitle

\begin{abstract}
In this paper, we introduce a neighbor embedding framework for manifold alignment. We demonstrate the efficacy of the framework using a manifold-aligned version of the uniform manifold approximation and projection algorithm. We show that our algorithm can learn an aligned manifold that is visually competitive to embedding of the whole dataset.
\end{abstract}

\section{Introduction}

Unsupervised clustering algorithms, like t-distributed stochastic neighbor embedding (t-SNE)~\citep{maaten2008visualizing} and uniform manifold approximation and projection (UMAP)~\citep{mcinnes2018umap}, map points in high-dimensional data space to a more visualizable low-dimensional space. They are growing in popularity, as the learning is unsupervised and the resulting clusters are often interpretable. However, the methods break down when points are taken from disparate datasets, as there is no link for pairwise comparison as the neighborhood graph networks are disconnected. The problem of mapping correlated points in unrelated ways, known as manifold alignment~\citep{ma2012manifold}, occurs whenever post-analysis data is added, e.g., for longitudinal studies, data compilations, and multi-modal measurements. It also arises with privacy and proprietary concerns, when source data cannot be transferred off-site. In each case, silos of datasets are created that cannot be part of a central representation learning scheme.

Previous attempts at manifold alignment have used a variety of methods, including semi-supervised learning~\citep{ham2005semisupervised}, spectral techniques~\citep{wang2009general}, and Procrustes analysis~\citep{wang2008manifold}. While the former uses known labels or bases for correlation, the latter is topological and integrates well with UMAP. Nevertheless, Procrustes transformations are not sufficient to align shared information among the datasets. This is because, Procrustes transformation is composed of various linear transforamtions which work on all the samples as an envelope of the data instead of individual samples. Here, we overcome this issue by iteratively optimizing the neighbor analysis for each dataset and jointly embedding the shared points. We mathematically characterize the new approach and experimentally validate it on several widely used datasets.

\section{Manifold-aligned neighbor embedding (MANE) framework}~\label{sec:man_alig_ne}

Assume the individual $n$-dimensional local datasets $\mathcal{D}^{(m)}=\{\mathbf{z}_i^{(m)}\}$, where $\mathbf{z}_i\in \mathbb{R}^n$ and $m=1,2,3,\dots,M$, cannot interact with each other. The index $i=1,2,3,\dots,N_m$ indexes each data point in the dataset. These datasets are local as in the datasets my be located in different physical systems that cannot interact with each other or hidden from each other. We construct a seeding dataset $\mathcal{D}^{(0)}=\{\mathbf{z}_i\}$, where $i=1,2,3,\dots,N_0$, in order to construct extended datasets $D^{(m)}=\mathcal{D}^{(0)} \cup \mathcal{D}^{(m)}=\{\mathbf{x}_i^{(m)}\}$, where $i=1,2,3,\dots,N_0,N_0+1,N_0+2,\dots,N_0+N_m$. Alternatively, $\mathcal{D}^{(0)}$ is the shared data among all the datasets $D^{(m)}$. For notational simplicity and without loss of generality, we assume that $\mathbf{x}_0^{(m)}=\mathbf{z}_0,\mathbf{x}_1^{(m)}=\mathbf{z}_1,\dots,\mathbf{x}_{N_0}^{(m)}=\mathbf{z}_N\in \mathcal{D}^{(0)}$ and $\mathbf{x}_{N_0+1}^{(m)},\mathbf{x}_{N_0+2}^{(m)},\dots,\mathbf{x}_{N_0+N_m}^{(m)}\in \mathcal{D}^{(m)}$. Now we can define a high-dimensional weighted graph $p^{(m)}$ for each of the datasets using pairwise metric $d_H(\cdot,\cdot)$. The entries of the adjacency matrix are given by
\begin{align}
    p^{(m)}_{i,j} = f_H(d_H(\mathbf{x}^{(m)}_i,\mathbf{x}^{(m)}_j),D^{(m)})
\end{align}
where $f_H(\cdot)$ is a function that describes the weighted relation between points $\mathbf{x}_i, \mathbf{x}_j\in D^{(m)}$ subject to the distance metric. In t-SNE the function $f_H(\cdot)$ is a Gaussian, whereas in UMAP it's k-nearest neighbor based affinity function. It is to be noted that, this construction of graph $p^{(m)}$ can be computed in the same location where dataset $\mathcal{D}^{(m)}$ is located. In what follows, the embedding algorithm or the global part of the algorithm effectively requires this adjacency matrix $p^{(m)}$ in order to compute the low-dimensional embedding, and any access to individual data points is not necessary.

We initialize the low-dimensional ($d$-dimensional) embedding $|D^{(m)}|=\{\mathbf{y}_i^{(m)}\}$ for each dataset, where $\mathbf{y}_i^{(m)} \in \mathbb{R}^d$ is the corresponding low-dimensional mapping of the high-dimensional point $\mathbf{x}_i^{(m)}\in D^{(m)}$. Now we can define a low-dimensional weighted graph for each of the datasets using the following pairwise relation
\begin{align}
    q^{(m)}_{i,j} = f_L(d_L(\mathbf{y}^{(m)}_i,\mathbf{y}^{(m)}_j),|D^{(m)}|)
\end{align}
where $f_L(\cdot)$ is a function of the weighted relation in the low dimension and $d_L(\cdot)$ is some pairwise metric. Typically, $d\ll n$. In t-SNE $f_L(\cdot)$ is the normalized Student's t-distribution function with one degree of freedom. UMAP skips the normalization step and uses a modified function similar to Student's t-distribution function of t-SNE's but has tunable parameters.

Finally the relation between the high-dimensional graphs and their joint low-dimensional embedding is established by optimizing the following problem
\begin{align}
    & \min_{|D^{(1)}|,\dots,|D^{(m)}|} \sum_{m} \sum_{i,j} l(p^{(m)}_{i,j},q^{(m)}_{i,j}) \nonumber \\
    & \text{s.t.} \nonumber \\
    & \mathbf{y}_i^{(0)} = \mathbf{y}_i^{(1)} = \dots = \mathbf{y}_i^{(M)}, \forall i = 1, 2, \dots N_0, \label{eq:man_align_loss}
\end{align}
where, $l(\cdot,\cdot)$ is the loss function. The constraint in Eq.~\ref{eq:man_align_loss} indicates that in MANE, the seeding dataset captures the shared manifold and works as anchor points, around which other data points are aligned in the low-dimensional embedding. To the best of our knolwedge this is the first time the constraint in Eq.~\ref{eq:man_align_loss} is being used for manifold alignment.

\section{Experiments}

In this section, we briefly describe the implementation of the framework in an algorithm and the evaluation metrics to validate our results. In the main text, we describe our results for Fashion-MNIST~\citep{xiao2017_online}. Results for additional datasets are described in the Appendix~\ref{appendix:results}.

This framework can be implemented using the principles of most modern dimensionality reduction methods. Our implementation of the framework is based on the UMAP algorithm. We follow the UMAP principles to build the high-dimensional graphs $p^{(m)}_{i,j}$, low-dimensional graphs $q^{(m)}_{i,j}$ for each dataset $D^{(m)}$. Then we define the loss function $l(p^{(m)}_{i,j},q^{(m)}_{i,j})$ to be the cross-entropy loss function. We optimize the embeddings of each dataset jointly using the negative sampling~\cite{mikolov2013distributed} approach. We sample one point from one of the datasets and optimize the loss function employing the constraint in Eq.~\ref{eq:man_align_loss}. More details of the implementation are provided in Appendix~\ref{appendix:implementation}. 

We compare our results to two other UMAP based methods: 1) individual UMAP: when the datasets are embedded individually and 2) aligned UMAP: a software package of UMAP that aligns the learned manifolds using Procrustes analysis, then optimizes each of the embeddings with a regularizer constraint on the shared points. For each comparison, the number of nearest neighbors, minimum distance parameter, and negative sampling rate have been set to 30, 0.1, and 1.0, respectively, for all embeddings. Aligned UMAP optimizes each dataset for 200 epochs individually and then optimizes the aligned embeddings for further 200 epochs. For UMAP implementation, the precomputed distance matrix has been used. Our MANE embeddings were obtained using 200 epochs and no pre-embeddings were required. We initialized the embeddings using the axes obtained from the principal component analysis (PCA) of the shared data. To compare the embeddings numerically, we use Procrustes distance ($d_p$) and trustworthiness ($T$)~\citep{venna2001neighborhood} metrics. Procrustes distance is the minimum euclidean distance between two datasets under translation, scaling, rotation, and reflection. Trustworthiness is a measure that gives a sense of how much of the nearest neighbors in the high dimension are preserved in the low dimension after embedding. More details of these metrics are provided in Appendix~\ref{appendix:metrics}. The PCA initializatin scheme can be used as a linear baseline, which has been further illustrated in Appendix~\ref{appendix:linear_baseline}.

\begin{figure} [t]
    \begin{center}
    \includegraphics[width=\linewidth]{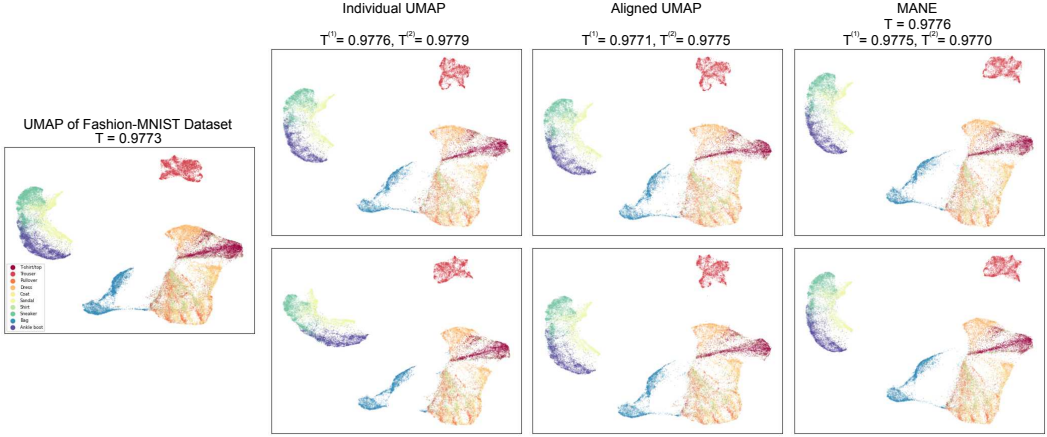}
    \end{center}
    \caption{Two-dimensional embedding of Fashion-MNIST data. (Left) UMAP embedding of 60,000 points ($T=0.9773$). (Right) Top row: embedding of $D^{(1)}$ and bottom row: embedding of $D^{(2)}$ for the individual UMAP, aligned UMAP, and MANE. Individual UMAPs naturally cannot align the manifolds which can be seen from misalignment of the large cluster consisting of images of ankle boot, sandal and sneaker in the two embeddings ($T^{(1)}=0.9776, T^{(2)}=0.9779$). Aligned UMAP ($T^{(1)}=0.9771, T^{(2)}=0.9775$) and MANE ($T=0.9769, T^{(1)}=0.9775, T^{(2)}=0.9770$) show very good alignment.}
    \label{fig:fmnist_embeddings}
\end{figure}

\begin{figure} [t]
    \begin{center}
    \includegraphics[width=\linewidth]{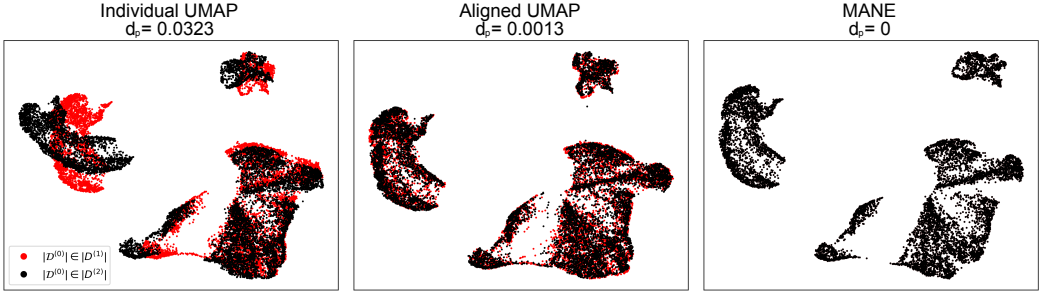}
    \end{center}
    \caption{Shared data points from the two-dimensional embedding of the Fashion-MNIST dataset of Figure~\ref{fig:fmnist_embeddings}. (Left) Individual UMAPs of sets $D^{(1)}$ and $D^{(2)}$ show that the shared information is not aligned between two datasets ($d_p=0.0323$). (Middle) Aligned UMAP shows close alignment between the shared points ($d_p=0.0013$). (Right) MANE shows best alignment between the shared points as ensured by the constraint in Eq~\ref{eq:man_align_loss} ($d_p=0$).}
    \label{fig:dodo_fmnist}
\end{figure}

The Fashion-MNIST dataset contains 70,000 gray-scale images of fashion items. The training data contains 60,000 images, which have been used in the experiments. First, we show results by randomly splitting the Fashion-MNIST dataset into three datasets $\mathcal{D}^{(m)}$, where $m=1,2,3$, where $\mathcal{D}^{(0)}$ contains 10,000 samples and the other splits contains 25,000 samples each. Then we construct $D^{(1)}=\mathcal{D}^{(0)}\cup \mathcal{D}^{(1)}$ and  $D^{(1)}=\mathcal{D}^{(0)}\cup \mathcal{D}^{(2)}$. The embeddings obtained using different schemes are shown in Figure~\ref{fig:fmnist_embeddings}. Since Individual UMAP and aligned UMAP provide separate embeddings for the shared points, we compute the trustworthiness metric of each dataset ($T^{(1)}$ and $T^{(2)}$). On the other hand, our method embeds both datasets into the same metric space and thus we can compute trustworthiness as if the whole Fashion-MNIST dataset has been embedded. Thus in addition to reporting $T^{(1)}$ and $T^{(2)}$, we also report trustworthiness of union of the two embeddings ($T$). We can observe from the figure that, all embeddings capture the general manifold of the Fashion-MNIST data as shown in the UMAP of the Fashion-MIST dataset. However, in individual UMAPs the embeddings $|D^{(1)}|$ and $|D^{(2)}|$ are not aligned to each other. This can be realized by observing the large cluster on the left involving the labels sneaker, ankle boot and sandal (colored teal, violet and greenish-yellow, respectively) which is oriented differently in the two embeddings indicating misalignment. Moreover, the other cluster on the lower right involving the labels t-shirt, pullover and dress (colored maroon, orange, and light green, respectively) is more compact in $|D^{(2)}|$ than it is in $|D^{(1)}|$. On the other hand, both aligned UMAP and MANE produce embeddings that are aligned to each other as the discrepencies described above are absent for these two. The trustworthiness of MANE is closer to the trustworthiness metric when the whole dataset has been embedded using UMAP. The trustworthiness metric $T^{(1)}$ and $T^{(2})$ are similar for all three metric, which indicates all of them have similar performance for embedding $D^{(1)}$ and $D^{(2)}$ whereas, the added benefit of aligned UMAP and MANE is the manifold alignment. While aligned UMAP uses regularization constraint for alignment (along with some costly pre-processing), MANE uses a hard constraint described in Eq~\ref{eq:man_align_loss}. This shows that jointly aligning and embedding using a hard constraint of Eq.~\ref{eq:man_align_loss} is a viable way to obtain aligned embedding. In Figure~\ref{fig:dodo_fmnist}, we examine how the shared points have been embedded. We can see that the shared points are not in agreement for individual UMAP and aligned UMAP, but they are in agreement for MANE.

\begin{figure} [t]
    \begin{center}
    \includegraphics[width=\linewidth]{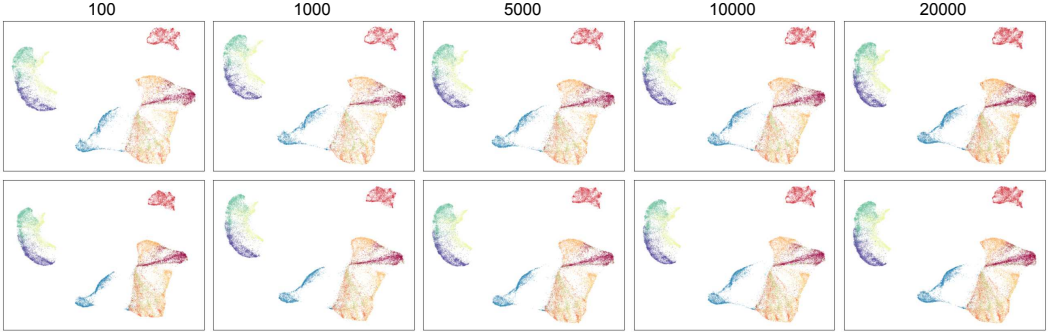}
    \end{center}
    \caption{MANE output of Fashion-MNIST data by setting  the number of shared points, $N_0$, to (from left to right) 100, 1000, 5000, 10000, and 20000 respectively. The data has been split into two datasets with $N_0$ shared points and then aligned using our implementation.  Top row: $|D^{(1)}|$ and bottom row: $|D^{(2)}|$. (from left to right) Trustworthiness values are $T=0.9722, 0.9754, 0.9765, 0.9769, 0.9771$. These embeddings show that the shared points have to sample the manifold enough to obtain better alignment.}
    \label{fig:vary_fmnist}
\end{figure}

\begin{figure} [t]
    \begin{center}
    \includegraphics[width=\linewidth]{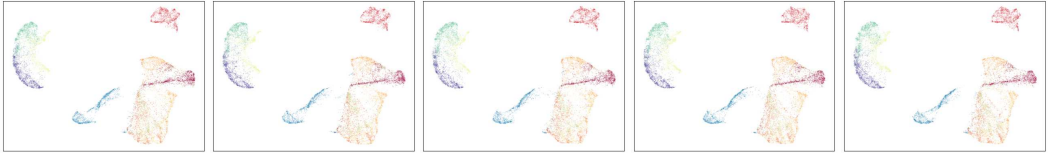}
    \end{center}
    \caption{MANE output of Fashion-MNIST data which is split into 5 datasets of 14400 data points. Each dataset shares 3000 points. The embeddings show good alignment ($T=0.9760$).}
    \label{fig:fmnist_5_split}
\end{figure}

Now we use the same scheme of the previous experiment but vary the value of $N_0$, the number of shared points, to see how much the shared information influences the alignment. We kept the total number of unique points to 60,000. The resultant embeddings are shown in Figure~\ref{fig:vary_fmnist}. It can be observed that except for $N_0=1000$ and $N_0=100$, all other embeddings are in good alignment. For $N_0=1000$ and $N_0=100$, the alignment is somewhat lost. For these embeddings, the large structures are in similar places having similar orientation, however, finer structures are not aligned. For example, for $N_0=1000$ and $N_0=100$, the larger cluster in the lower right is compact for $|D^{(2)}|$ compared to that of $|D^{(1)}|$. For $N_0 = 100$, the spike like structure in the cluster of sandal (colored greenish-yellow in the left of the figure) is different for $|D^{(1)}|$ and $|D^{(2)}|$. This is due to the fact that the shared points work as anchors around which the manifold of both datasets is arranged. However, with merely 1000 or 100 points, the shared points cannot sample enough of the manifold.

Finally, we look into a case where more than two datasets are involved. We split the Fashion-MNIST data into 5 datasets of equal size (14400 data points each). The number of shared points is 3000. The resultant aligned embeddings are shown in Figure~\ref{fig:fmnist_5_split}. The figures show that all 5 datasets are aligned.

\section{Conclusions and Future Work}

In this paper, we introduce MANE, a neighbor embedding approach for aligning the manifold of multiple datasets that use the same underlying manifold. We modeled the problem in terms of shared data points among multiple datasets. We implemented our approach using UMAP principles. The efficacy of our method in aligning datasets is demonstrated by several experiments. We showed the shared points can be in perfect alignment yet produce embeddings that are comparable to the embedding of the combined dataset. We compared our method with aligned UMAP in which the shared data are not in perfect alignment. Overall, MANE produce emeddings that are comparable to mebedding of combined dataset, and provides perfect alignment for the shared points. This method can be used in practice where shared data is known or a seeding dataset is available.

In the future, a more interesting demonstration would be in a case where two datasets are of different modalities (say x-ray image and its corresponding report or two different medical image modalities). The dataset $\mathcal{D}^{(0)}$ contains pairs of x-ray image and reports, whereas, $\mathcal{D}^{(1)}$ and $\mathcal{D}^{(2)}$ are datasets of x-rays, and reports, respectively. Thus, using manifold-aligned neighbor embedding one can jointly obtain the embedding of x-rays and reports, which will be of great importance in medical AI.

\section*{Acknowledgement}
This material is based upon work supported by the Air Force Office of Scientific Research (AFOSR) under Grant FA9550-18-1-0219 and by the Defense Advanced Research Projects Agency (DARPA) under Agreement No. HR00112090123. The views and conclusions contained herein are those of the authors and should not be interpreted as necessarily representing the official policies or endorsements, either expressed or implied, of DARPA, AFOSR, or the U.S. Government.

\section*{Code and Data Availability}

The codebase of the paper can be obtained from \url{https://github.com/tariqul-islam/mane_paper}.

All the data used in this paper are publicly availble. \\
Fashion-MNIST is available at \url{https://github.com/zalandoresearch/fashion-mnist}. \\
MNIST is available at \url{http://yann.lecun.com/exdb/mnist/}.\\
Swiss roll data can be generated from \url{https://scikit-learn.org/stable/modules/generated/sklearn.datasets.make_swiss_roll.html}. \\
Single cell transcriptomes data is available at \url{https://github.com/biolab/tsne-embedding}.

\bibliography{biblio_manifold}
\bibliographystyle{iclr2022_workshop}

\clearpage

\appendix

\section{Implementation}~\label{appendix:implementation}

In this paper, we show an implementation of the general framework described in Section~\ref{sec:man_alig_ne} using UMAP principles. We define the high-dimensional weighted relation for each dataset~$D^{(m)}$ by
\begin{align}
    p_{i|j}^{(m)} =
        \begin{cases}
            \exp{\left(-\frac{d(\mathbf{x}^{(m)}_i,\mathbf{x}^{(m)}_j)-\rho^{(m)}_i}{\sigma^{(m)}_i}\right)} & \text{if } x^{(m)}_j\in \text{KNN}(\mathbf{x}^{(m)}_i,k) \\
            0 & \text{otherwise}
        \end{cases} \label{eq:UMAP_HIGH_DIM}
\end{align}
where $\text{KNN}(\mathbf{x}^{(m)}_i,k)$ is the set of $k$-nearest neighbors in dataset $D^{(m)}$ of the point $\mathbf{x}^{(m)}_i$, $\rho^{(m)}_i = \min_{\mathbf{x}^{(m)}_j\in \text{KNN}(\mathbf{x}^{(m)}_i,k)} d(\mathbf{x}^{(m)}_i,\mathbf{x}^{(m)}_j)$ and $\sigma^{(m)}_i$ is a scaling parameter set such that $\sum_j  p^{(m)}_{i|j} = \log_2(k)$. The parameter $\rho^{(m)}_i$ ensures that the point $\mathbf{x}^{(m)}_i$ has strong connectivity ($p^{(m)}_{i|j}=1$) to at least one of the nearest neighbors and the scaling parameter $\sigma^{(m)}_i$ ensures the uniform manifold approximation. The adjacency matrix $p^{(m)}$ of the dataset $D^{(m)}$ in the high-dimension is then obtained by (the probabilistic t-conorm), 
\begin{align}
    p^{(m)}_{ij} = p^{(m)}_{i|j} + p^{(m)}_{j|i} - p^{(m)}_{i|j} \times p^{(m)}_{j|i}
\end{align}

We define the low-dimensional weighted relation by a differentiable function
\begin{align}
    q^{(m)}_{ij} = 
    \frac{1}{1+a(||\mathbf{y}^{(m)}_i-\mathbf{y}^{(m)}_j||_2^2)^b} \label{eq:low_dim_umap}
\end{align}
where the parameters $a$ and $b$ determine how crowded the low-dimensional points become after embedding. These two parameters are chosen by fitting $q_{ij}$ to 
\begin{align}
    \Psi(\mathbf{y}^{(m)}_i,\mathbf{y}^{(m)}_j) 
    = 
    \begin{cases}
    1 & \text{if } ||\mathbf{y}_i-\mathbf{y}^{(m)}_j||_2<d_m \\
    \exp(-(||\mathbf{y}^{(m)}_i-\mathbf{y}_j||_2 - d_m)) & \text{otherwise}
    \end{cases}
\end{align}
where $d_m$ is a user-defined parameter that regulates the minimum distance between two low-dimensional points.

Finally, we define the loss function
\begin{align}
    l(p^{(m)}_{i,j},q^{(m)}_{i,j}) 
    = p^{(m)}_{ij} \log\left( \frac{p^{(m)}_{ij}}{q^{(m)}_{ij}} \right) +
    \left(1-p^{(m)}_{ij}\right) \log\left( \frac{1-p^{(m)}_{ij}} {1-q^{(m)}_{ij}} \right). \label{eq:crs_loss_fun}
\end{align}
The first term in the loss function provides the attractive force to bring similar points closer together, and the second term provides the repulsive force that enables far apart points in the high dimension to stay far apart in the low dimension.

To optimize this loss function, we employed the negative sampling~\citep{mikolov2013distributed} approach that was used in the original UMAP implementation. In each optimization step, we sample one positive edge from one of the graphs and apply the attractive force to it. Then we sample negative edges and apply the repulsive forces to these edges. For a positive edge which is part of the seeding dataset, we apply the constraint in Eq.~\ref{eq:man_align_loss} so that the embedding of the shared point is same across all datasets. In the code, we ensure this by having a shared memory for these shared embedded points.

\clearpage

\section{Evaluation Metrics}\label{appendix:metrics}
\subsection{Trustworthiness}

The trustworthiness metric measures how much of the local structure has been preserved after dimensionality reduction by
\begin{align}
    {\tiny 
    T = 1 - 
\frac{2}{nk(2n-3k-1)} \sum_{i=1}^n \sum_{y_j\in \text{KNN}(y_i,k)} \max(0,r(i,j)-k)
    }%
\end{align}
where $KNN(y_i,k)$ is the $k$-NN graph in the embedding space and $r(i,j)$ is the rank of $x_j$ in the high-dimensional $k$-NN graph. Usually, the value of $k$ is set to 5, which considers only 5 nearest neighbors in the low-dimensional embedding.  

\subsection{Procrustes Distance}

Procrustes distance is the minimum distance between two sets of points under scaling, translation, reflection, and rotation. Functionally, for two datasets $\{\mathbf{x}_i\}$ and $\{\mathbf{y}_i\}$ we want to scale, translate and rotate one of the datasets in such a way that the two datasets are in maximum alignment. Assuming $\{\mathbf{y}'_i\}$ is a transformation of $\{\mathbf{y}_i\}$ that achieves the desired goal. Then Procrustes distance is given by
\begin{align}
    d_p = \sqrt{\sum_i ||\mathbf{x}_i-\mathbf{y}'_i||^2}
\end{align}

\clearpage

\section{Additional Results}~\label{appendix:results}

In this section, we describe the results of some additional datasets.

\subsection{Swiss Roll}

\begin{figure} [h]
    \begin{center}
    \includegraphics[width=\linewidth]{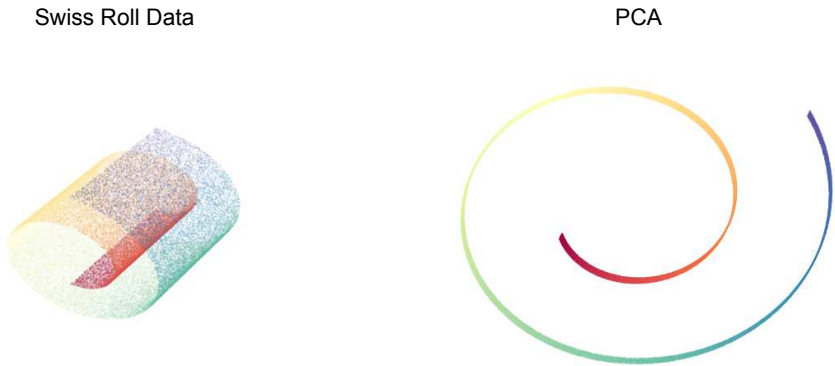}
    \end{center}
    \caption{60,000 samples from the Swiss roll function. (Left) 3D plot of the data. (Right) PCA projection of the data.}
    \label{fig:swiss_data}
\end{figure}

\begin{figure} [h]
    \begin{center}
    \includegraphics[width=\linewidth]{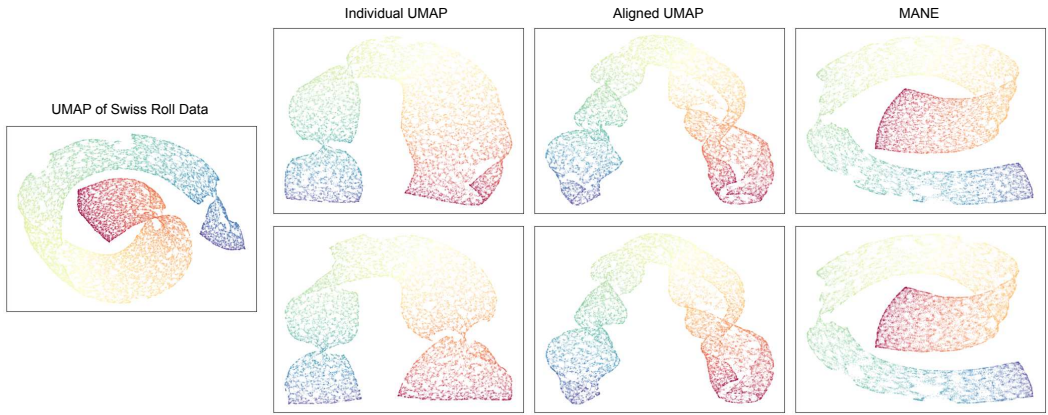}
    \end{center}
    \caption{Two dimensional embedding of Swiss roll data. (Left) UMAP embedding (PCA initialized) of 60,000 points ($T=0.9983$). (Right) Top row: $|D^{(1)}|$ and bottom row: $|D^{(2)}|$ for the individual UMAP, aligned UMAP, and MANE. Individual UMAPs naturally cannot align the manifolds ($T^{(1)}=0.9991, T^{(2)}=0.9990$). Aligned UMAP ($T^{(1)}=0.9894, T^{(2)}=0.9874$) and MANE ($T=0.9978, T^{(1)}=0.9975, T^{(2)}=0.9981$) show very good alignment.}
    \label{fig:swiss_embeddings}
\end{figure}

\begin{figure} [h]
    \begin{center}
    \includegraphics[width=\linewidth]{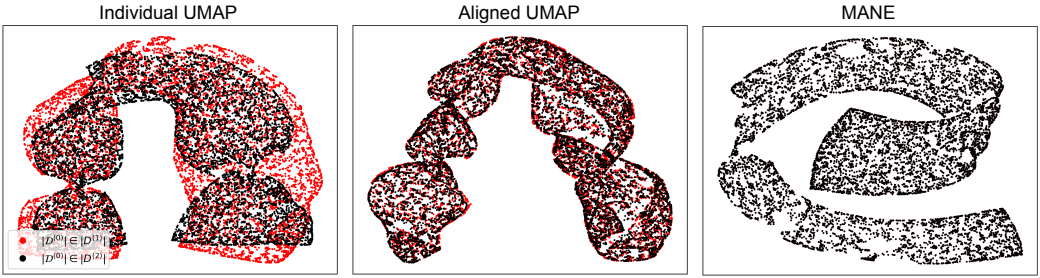}
    \end{center}
    \caption{Shared data points from the two-dimensional embedding of the Swiss roll data of Figure~\ref{fig:swiss_embeddings}. (Left) Individual UMAPs of sets $D^{(1)}$ and $D^{(2)}$ show that the shared information is not aligned between two datasets ($d_p=0.4191$). The points are superimposed here after Procrustes analysis. (Middle) Aligned UMAP shows close alignment between the shared points ($d_p=0.0221$). (Right) MANE shows best alignment ($d_p=0$).}
    \label{fig:dodo_swiss}
\end{figure}

\begin{figure} [h]
    \begin{center}
    \includegraphics[width=\linewidth]{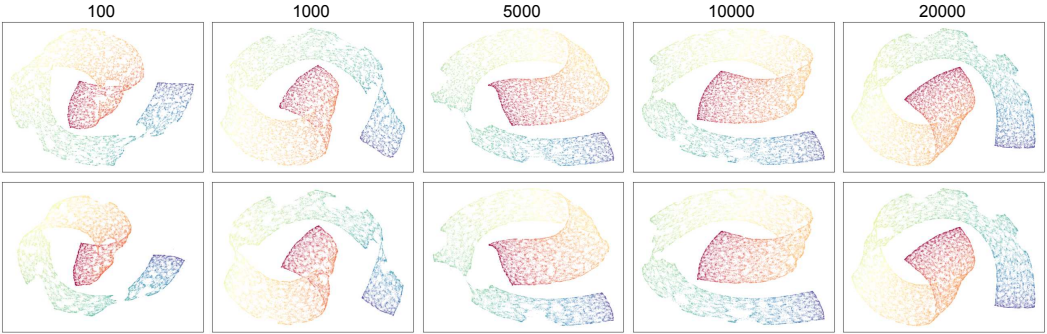}
    \end{center}
    \caption{Manifold-aligned neighbor embedding for Swiss roll data for varying the number of shared points ($N_0$). The data has been split into two datasets with $N_0$ shared points and then aligned using our implementation (trustworthiness from left to right $T=0.9943, 0.9952, 0.9978, 0.9978, 0.9950$).  Top row: $|D^{(1)}|$ and bottom row: $|D^{(2)}|$. Similar to Fashion-MNIST, the alignment is not optimal for $N_0=100$. Which can be seen from the purple colored tail that is narrower in $|D^{(2)}|$. As $N_0$ increases the alignmnets becomes better (which can be observed from the increasig trustworthiness value as well).}
    \label{fig:vary_swiss}
\end{figure}

\begin{figure} [h]
    \begin{center}
    \includegraphics[width=\linewidth]{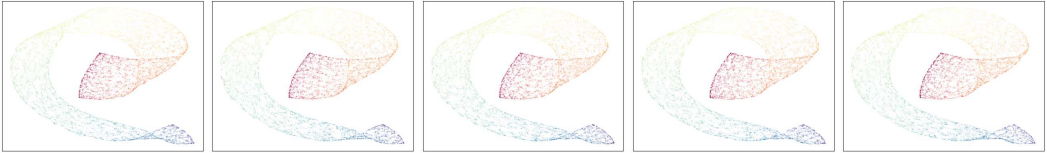}
    \end{center}
    \caption{MANE output of Swiss roll data which is split into 5 datasets of 14400 data points. Each datasets share 3000 points. The embeddings show good alignment ($T=0.9885$).}
    \label{fig:swiss_5_split}
\end{figure}

The Swiss roll~\citep{roweis2000nonlinear} is a two-dimensional manifold in a 3D space. Figure~\ref{fig:swiss_data} shows 60,000 samples from the manifold. The data are colored according to their position in the manifold to understand their apparent location in the original space. The experiments performed are similar to that of Fashion-MNIST data that was performed in the main text. Figures~\ref{fig:swiss_embeddings}-\ref{fig:swiss_5_split} describe the results.

\clearpage

\subsection{MNIST Dataset}

\begin{figure} [h]
    \begin{center}
    \includegraphics[width=\linewidth]{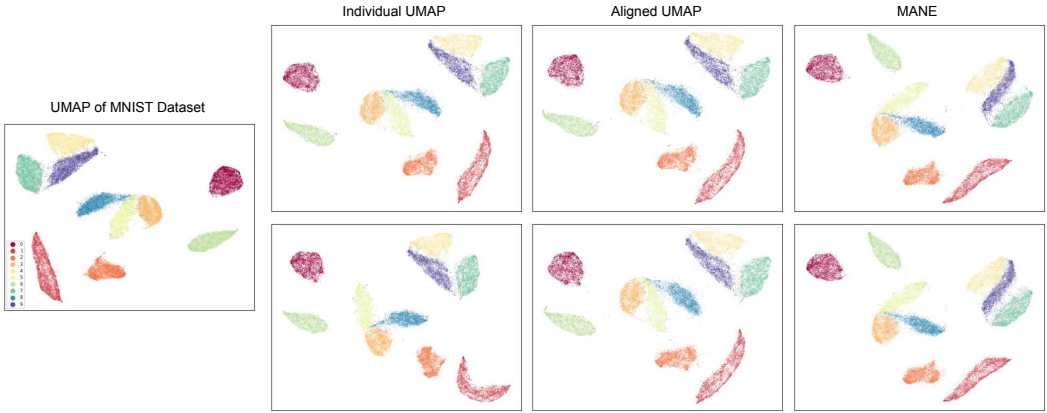}
    \end{center}
    \caption{Embedding of MNIST dataset. (Left) UMAP embedding of 60,000 points ($T=0.9573$). (Right) Top row: embedding of $D^{(1)}$ and bottom row: embedding of $D^{(2)}$ for the individual UMAP, aligned UMAP, and our manifold-aligned neighbor embedding. Individual UMAPs naturally cannot align the manifolds ($T^{(1)}=0.9570, T^{(2)}=0.9582$). For example, the red clusters involding 0 have different shapes in $|D^{(1)}|$ and $|D^{21)}|$. Orintation of other clusters are also different. Aligned UMAP ($T^{(1)}=0.9568, T^{(2)}=0.9570$) and MANE ($T=0.9531, T^{(1)}=0.9536, T^{(2)}=0.9542$) show very good alignment.}
    \label{fig:mnist_embeddings}
\end{figure}

\begin{figure} [h]
    \begin{center}
    \includegraphics[width=\linewidth]{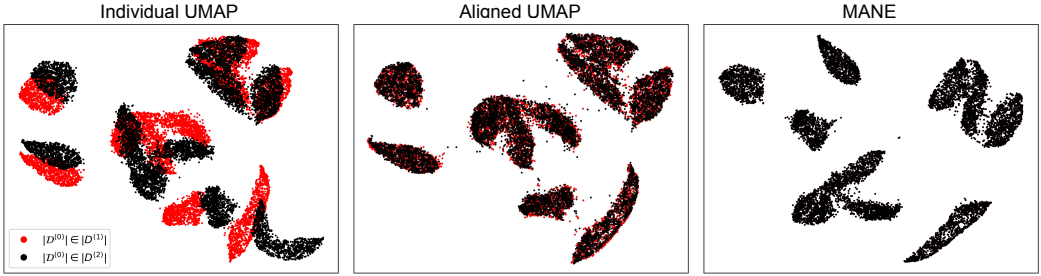}
    \end{center}
    \caption{Shared data points from the two-dimensional embedding of the Fashion-MNIST dataset of Figure~\ref{fig:fmnist_embeddings}. (Left) Individual UMAPs of sets $D^{(1)}$ and $D^{(2)}$ show that the shared information is not aligned between two datasets ($d_p=0.3458$). (Middle) Aligned UMAP shows close alignment between the shared points ($d_p=0.0421$). (Right) MANE shows best alignment ($d_p=0$).}
    \label{fig:dodo_mnist}
\end{figure}

\begin{figure} [h]
    \begin{center}
    \includegraphics[width=\linewidth]{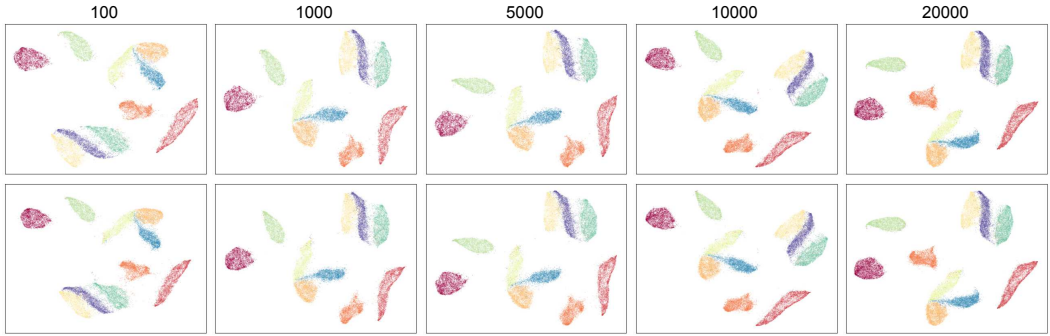}
    \end{center}
    \caption{Manifold-aligned neighbor embedding for MNIST dataset for varying the number of shared points ($N_0$). The data has been split into two datasets with $N_0$ shared points and then aligned using our implementation (trustworthiness from left to right $T= 0.93747, 0.9503 0.9514, 0.9531, 0.9543$). Top row: $|D^{(1)}|$ and bottom row: $|D^{(2)}|$. 
    These embeddings show that the shared points have to sample the manifold enough to obtain better alignment.}
    \label{fig:vary_mnist}
\end{figure}

\begin{figure} [h]
    \begin{center}
    \includegraphics[width=\linewidth]{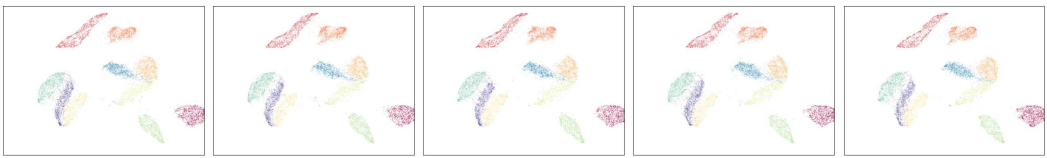}
    \end{center}
    \caption{MANE output of MNIST data which is split into 5 datasets of 14400 data points. Each datasets share 3000 points. The embeddings show good alignment ($T=0.9500$).}
    \label{fig:mnist_5_split}
\end{figure}

MNIST~\citep{lecun1998gradient} is a classic dataset used in many machine learning research. The dataset has the same structure as the Fashion-MNIST dataset. There are 60,000 training data and 10,000 test data consisting of 10 categorical classes. Our results here involve only the training data. The results are compiled in Figures~\ref{fig:mnist_embeddings}-\ref{fig:mnist_5_split}.

\clearpage

\subsection{Single-cell transcriptomes}

\begin{figure} [h]
    \begin{center}
    \includegraphics[width=\linewidth]{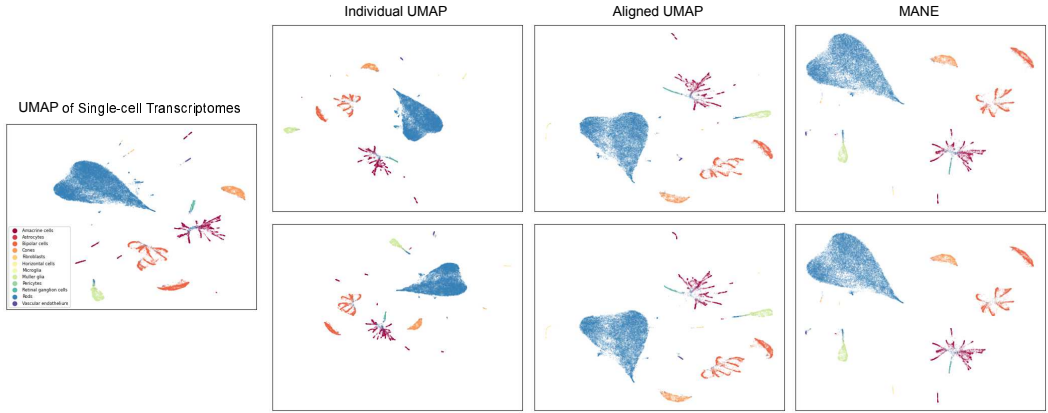}
    \end{center}
    \caption{Embedding of single-cell transcriptomes. (Left) UMAP embedding of 44,808 points ($T=0.9495$). (Right) Top row: embedding of $D^{(1)}$ and bottom row: embedding of $D^{(2)}$ for the individual UMAP, aligned UMAP, and our manifold-aligned neighbor embedding. Individual UMAPs naturally cannot align the manifolds ($T^{(1)}=0.9485, T^{(2)}=0.9488$). This can be observed by looking at the cluster involving bipolar cells (colored orange) which is orinted similarly in $|D^{(1)}|$ and $|D^{(2)}|$. However, the rest of the clusters are orinted differently and thus misaligned. Aligned UMAP ($T^{(1)}=0.9478, T^{(2)}=0.9482$) and MANE ($T=0.9446, T^{(1)}=0.9459, T^{(2)}=0.9443$) show very good alignment.}
    \label{fig:macosko_embeddings}
\end{figure}

\begin{figure} [h]
    \begin{center}
    \includegraphics[width=\linewidth]{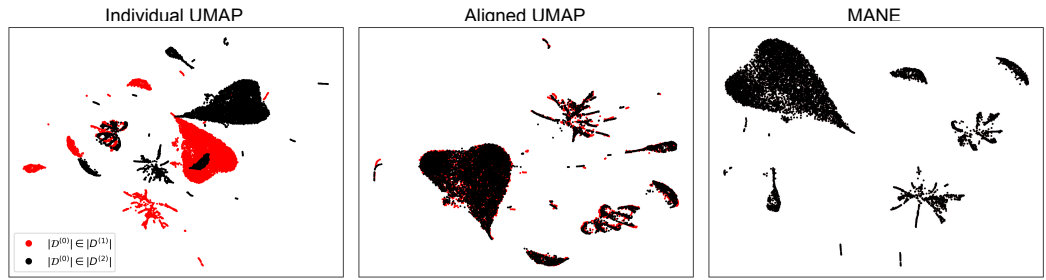}
    \end{center}
    \caption{Shared data points from the two-dimensional embedding of the single-cell transcriptomes of Figure~\ref{fig:fmnist_embeddings}. (Left) Individual UMAPs of sets $D^{(1)}$ and $D^{(2)}$ show that the shared information is not aligned between two datasets ($d_p=0.8288$). (Middle) Aligned UMAP shows close alignment between the shared points ($d_p=0.0488$). (Right) MANE shows best alignment ($d_p=0$).}
    \label{fig:dodo_macosko}
\end{figure}

\begin{figure} [h]
    \begin{center}
    \includegraphics[width=\linewidth]{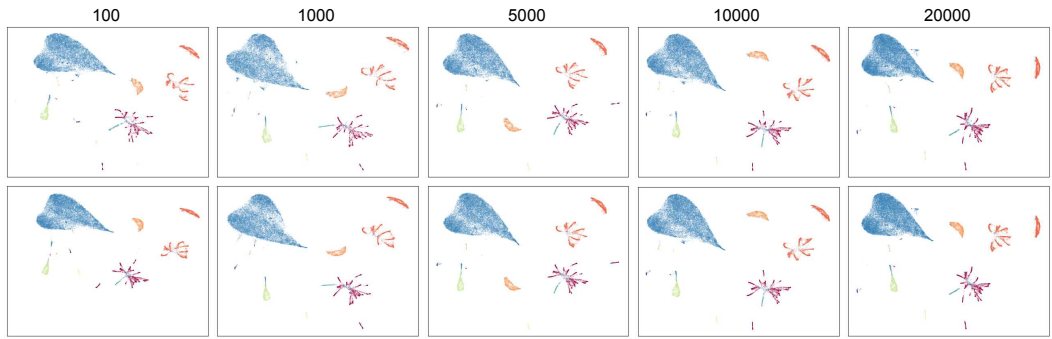}
    \end{center}
    \caption{MANE output of the single-cell transcriptomes by varying the number of shared points ($N_0$). The data has been split into two datasets with $N_0$ shared points and then aligned using our implementation.  Top row: $|D^{(1)}|$ and bottom row: $|D^{(2)}|$. Trustworthiness metrics from left to right are $0.9402$, $0.9431$, $0.9439$, $0.9446$, and $0.9469$, respectively. Due to the large number of small clusters in this dataset a small number of shared poaints may not be able to sample the manifold. It is evident from the figure that, for $N_0=100$ and $N_0=1000$, similar clusters in $|D^{(1)}|$ and $|D^{(2)}|$ do not cossespond to each other. For example, the cluster involving bipolar cells (orange octopus like structure) in the two embeddings are oriented differently for $N_0=100$. As $N_0$ is increased the alignment improves as the shared information can sample more of the manifold. }
    \label{fig:vary_macosko}
\end{figure}

\begin{figure} [h]
    \begin{center}
    \includegraphics[width=\linewidth]{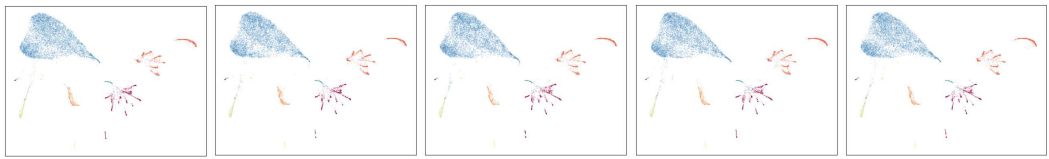}
    \end{center}
    \caption{MANE output ($T=0.9360$) of the single cell transcriptomes which is split into 5 datasets of 8360 data points. Number of shared points is 3000. Overall all 5 datasets are aligned to each other.}
    \label{fig:macosko_5_split}
\end{figure}

This dataset was compiled by ~\cite{macosko2015highly}. This includes 44,808 samples of single-cell transcriptomes obtained from the mouse retina. Results are compiled in Figures~\ref{fig:macosko_embeddings}-\ref{fig:macosko_5_split}. In figure~\ref{fig:macosko_embeddings} we split the data into two sets with 10,000 shared points. 

\clearpage

\section{Comparison to principal component analysis}~\label{appendix:linear_baseline}

In this section, we describe a linear baseline using PCA. In this scheme, we obtain the PCA axes by performing PCA on $\mathcal{D}^{(0)}$. Using the computed axes, we project each of the datasets, $D^{(i)}$ and obtain the low dimensional embeddings. This scheme, similar to MANE, also ensures that the Procrustes distance is 0 for the shared points.

\begin{figure} [h]
    \begin{center}
    \includegraphics[width=0.8\linewidth]{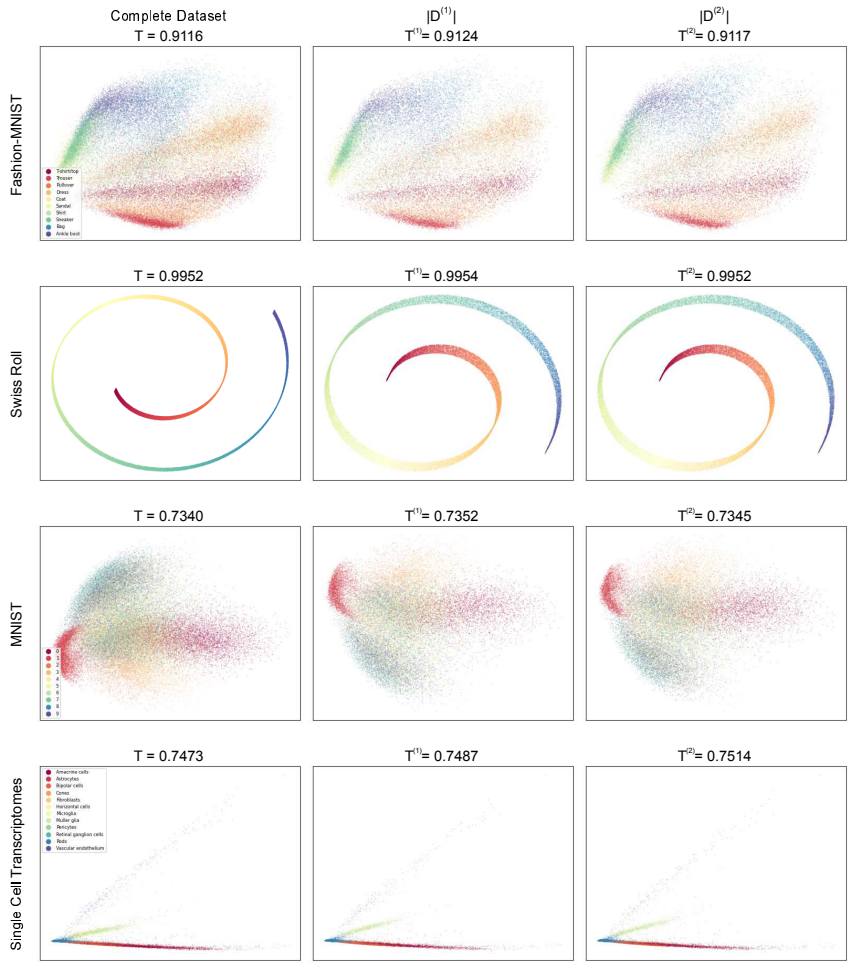}
    \end{center}
    \caption{Linear baseline using PCA. The trustworhiness scores are lower than that of the nonlinear embeddings.}
    \label{fig:PCA_figure}
\end{figure}

In the experiments, we use the same datasets we used in the previous sections. The number of shared points $N_0$ is set to 10,000 and rest of the data is split into two datasets to obtain $D^{(1)}$ and $D^{(2)}$. The results are shown in Figure~\ref{fig:PCA_figure}. We can observe that, while the data are aligned to each other, the trustworthiness scores are significanly lower for the linear baseline except for Swiss roll. Swiss roll data is a 2-dimensional manifold in 3-dimension which is easier to project in 2-dimension using PCA axes compared to the rest of the data.

\end{document}